\documentclass[10pt,twocolumn,letterpaper]{article}

\usepackage{iccv}
\usepackage{times}
\usepackage{epsfig}
\usepackage{graphicx}
\usepackage{amsmath}
\usepackage{graphics}
\usepackage{multirow}
\graphicspath{./figures/}
\usepackage{amssymb}
\usepackage{algorithm}
\usepackage{algpseudocode}
\usepackage{caption}
\usepackage{subcaption}
\usepackage[accsupp]{axessibility} 

\usepackage[breaklinks=true,bookmarks=false]{hyperref}

\iccvfinalcopy 

\def\iccvPaperID{4072} 

\ificcvfinal\fi

\begin{document}

\title{ReST: A Reconfigurable Spatial-Temporal Graph Model for Multi-Camera Multi-Object Tracking}


\author{Cheng-Che Cheng$^1$ \quad Min-Xuan Qiu$^1$ \quad Chen-Kuo Chiang$^2$ \quad Shang-Hong Lai$^1$\\
$^1$National Tsing Hua University, Taiwan \quad $^2$National Chung Cheng University, Taiwan\\
{\tt\small \{chengche6230, maisiechiu\}@gapp.nthu.edu.tw \quad ckchiang@cs.ccu.edu.tw \quad lai@cs.nthu.edu.tw}
}

\maketitle
\ificcvfinal\thispagestyle{empty}\fi

\begin{abstract}
   Multi-Camera Multi-Object Tracking (MC-MOT) utilizes information from multiple views to better handle problems with occlusion and crowded scenes. Recently, the use of graph-based approaches to solve tracking problems has become very popular. However, many current graph-based methods do not effectively utilize information regarding spatial and temporal consistency. Instead, they rely on single-camera trackers as input, which are prone to fragmentation and ID switch errors. In this paper, we propose a novel reconfigurable graph model that first associates all detected objects across cameras spatially before reconfiguring it into a temporal graph for Temporal Association. This two-stage association approach enables us to extract robust spatial and temporal-aware features and address the problem with fragmented tracklets. Furthermore, our model is designed for online tracking, making it suitable for real-world applications. Experimental results show that the proposed graph model is able to extract more discriminating features for object tracking, and our model achieves state-of-the-art performance on several public datasets. Code is available at \url{https://github.com/chengche6230/ReST}.
\end{abstract}

\vspace{-1em}
\section{Introduction}

\begin{figure}[t]
\begin{center}
\begin{subfigure}[b]{1\linewidth}
         \centering
         \includegraphics[width=0.9\linewidth]{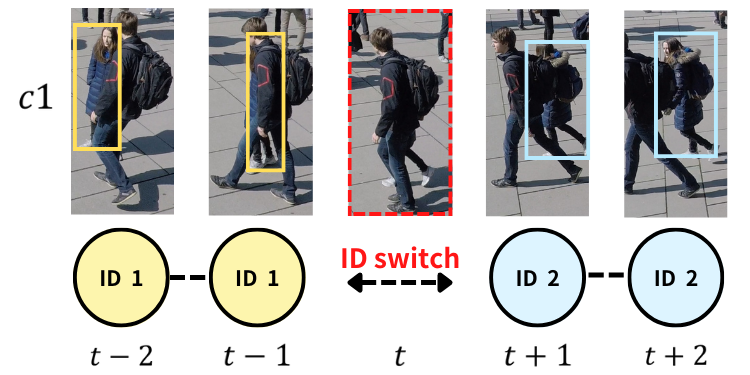}
         \caption{Single-camera tracker}
         \label{fig:fig1-a}
\end{subfigure}

\begin{subfigure}[b]{1\linewidth}
         \centering
         \includegraphics[width=1\linewidth]{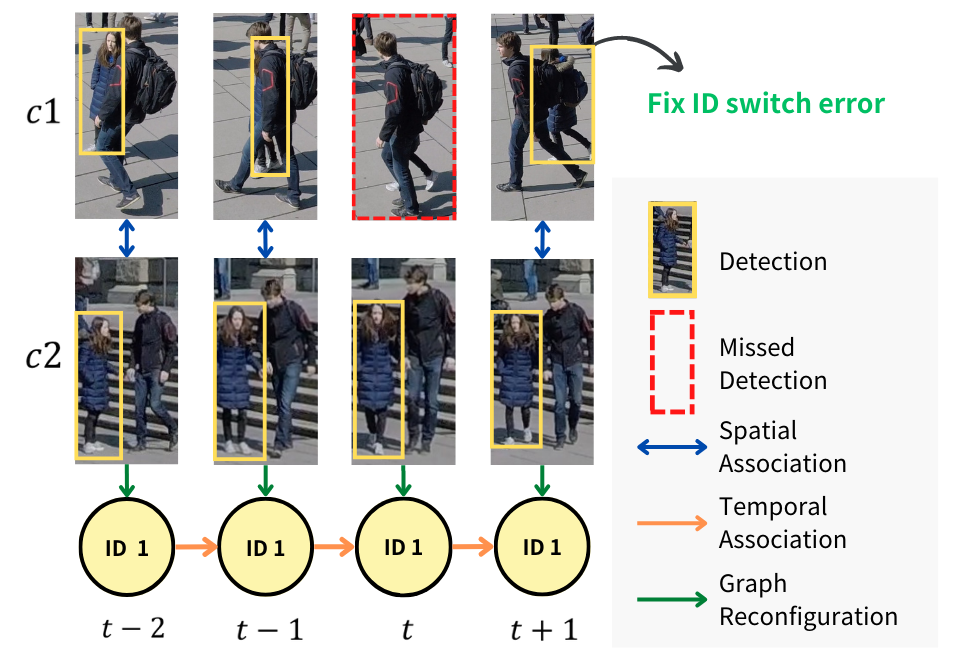}
         \caption{Our ReST tracker}
         \label{fig:fig1-b}
\end{subfigure}
\end{center}
   \caption{Example of handling object occlusion for MC-MOT. (a) When occlusion occurs at time $t$ (red dotted box), the single-camera tracker generates fragmented tracklets and causes ID switch errors. (b) Our ReST tracker corrects object ID in $c1$ via Spatial and Temporal Association, by leveraging spatial and temporal consistency.}
\vspace{-1.5em}
\label{fig:fig1}
\end{figure}

Multi-Object Tracking (MOT) is an important task in computer vision, which involves object detection and tracking multiple objects over time in an image sequence. It can be applied to several real-world scenarios, such as video surveillance, autonomous vehicles, and sports analysis. Despite numerous research methods proposed for MOT, the problem of fragmented tracklets or ID switching caused by frequent occlusion in crowded scenes remains a major challenge.  One potential solution is to track objects under a multi-camera setting, which is called a Multi-Camera Multi-Object Tracking (MC-MOT) task. By leveraging information from multiple cameras, occluded objects in one view may become clearly visible in another view, allowing for more accurate object tracking results.

Most of tracking-by-detection paradigms \cite{tracking-by-detection} adopt Kalman filter \cite{kalman} in the data association stage. It serves as a motion model, predicting the next possible position and matching with previous detection. However, such method is usually deterministic and cannot adapt to the dynamically changing environment. In addition, the tracking results are difficult to achieve globally optimal, since the illumination, relative geometry distance, or sampling rate varies from dataset to dataset, which is common in real-world scenarios. Accordingly, there is another fashion reformulating the association problem into link prediction on graph \cite{neuralsolver,polarmot,LMGP,DyGLIP}. It allows a trainable model to determine how strong the connection is between two detections. Thus, objects can be dynamically associated depending on environmental conditions.

However, there still remains some issues in current graph-based models for MC-MOT.
First of all, many approaches rely on single-camera tracker to generate the initial tracklets \cite{TRACTA,LMGP,DyGLIP,HCT}. Although many methods have been proposed to refine tracklets, tracking errors in single-view are often left unaddressed. Additionally, these methods do not fully leverage the rich spatial and temporal information that is crucial for MC-MOT task. Recently, spatial-temporal models have been employed to learn representative features for tracklets. However, the resulting graphs are usually complex and hard to optimize.

In this paper, we propose a novel Reconfigurable Spatial-Temporal graph model (ReST) for MC-MOT to overcome the problems mentioned above. The MC-MOT problem is re-formulated as two sub-tasks, Spatial Association and Temporal Association, in our approach. In Spatial Association, it focuses on matching objects across different views. Temporal Association exploits temporal information, such as speed and time, to build temporal graph which associates objects across frames. By splitting the problem into two sub-tasks, spatial and temporal consistency can be individually optimized to achieve better tracking results. In addition, the graph model becomes smaller and easy to optimize. To bridge two association stages, Graph Reconfiguration module is proposed to aggregate information from spatial and temporal graph models. The merits of involving graph reconfiguration are two-fold. Firstly, when the nodes of the same object are merged, the reconfigured graph becomes very compact. Secondly, the refinement of the graph model can be iteratively performed in each reconfiguration step during inference, leading to more representative feature extraction and better tracking results. As depicted in Figure~\ref{fig:fig1-a}, when the girl is occluded, fragmented tracklets are produced, causing the ID switch problem. In Figure~\ref{fig:fig1-b}, correct object ID can be retained by employing spatial and temporal consistency via Spatial Association, Temporal Association, and Graph Reconfiguration modules.

The proposed graph model is called reconfigurable because the vertex set and edge set of spatial and temporal graphs are reconfigured to construct a new graph at each time. Thus, it tends to adapt to dynamic scenes. Unlike existing methods, our model does not rely on the results from single-camera tracker. The tracking and association of the detected objects is accomplished through iteratively constructing spatial and temporal graphs. Our model is designed for online object tracking since it does not use or rely on any information from future frames.

\vspace{-1em}
\paragraph{Contributions}
Our contributions can be summarized as follows. 1) The Multi-Camera Multi-Object Tracking problem is formulated as two sub-tasks in the proposed graph model, Spatial Association and Temporal Association. This enables the employment of spatial and temporal consistency and better model optimization. 2) Graph Reconfiguration module is proposed to leverage tracking results from two stages. This makes the object tracking apt to dynamic scene changes and online tracking scenarios. 3) Experimental results demonstrate that our model achieves state-of-the-art performance on Wildtrack and competitive results on other benchmark datasets.

\section{Related Work}

In recent years, a number of research works have focused on single-camera MOT. For example, \cite{ocsort,TransMOT,MHT,bytetrack} focus on improving data association and precisely extracting motion. \cite{bnw,famnet,motionmodel,JDE,trainmot,point} unify the object detection and association stage into an end-to-end model.
Recently, MC-MOT has received significant attention and grown increasingly \cite{wildtrack,TRACTA,HJMV,LMGP,GLMB,DyGLIP,HCT,STP,tglimmpse}.
Although it contains more spatial-temporal information than single-camera tracking problem, the MC-MOT problem still presents several challenges that must be overcome,  including varying environmental conditions and the lack of integration of spatial-temporal information.

\begin{figure*}
\begin{center}
\includegraphics[width=1\linewidth]{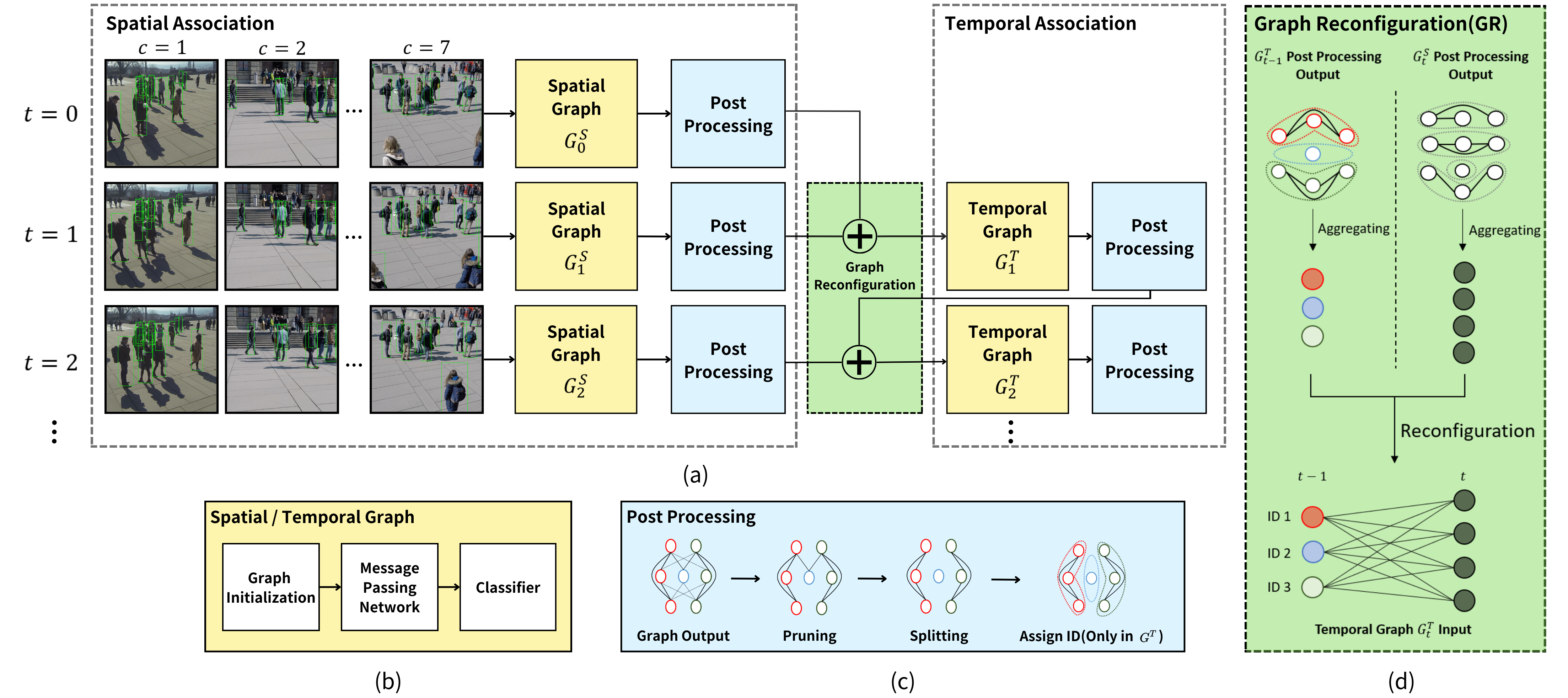}
\end{center}
\caption{Proposed ReST framework. (a) Inference architecture: given input detection from all views at time $t$, Spatial Association is performed and followed by Graph Reconfiguration and Temporal Association. (b) Graph model learning: both $G^S$ and $G^T$ are trained following the steps: initialization, message passing network, and edge classification. (c) Post-processing module: output a refined graph by two schemes: pruning and splitting. (d) Graph Reconfiguration: aggregate graph nodes from $G^T_{t-1}$ and $G^S_t$ and reconfigure a new temporal graph $G^T_t$.}
\vspace{-1.5em}
\label{fig:framework}
\end{figure*}

\vspace{-1em}
\paragraph{Spatial-Temporal Representation Learning}
The spatial and temporal feature is a key factor in motion-related areas, such as human pose estimation, and MOT. \cite{tglimmpse} sets up an occupancy map to fuse cross-view spatial correlations, followed by a Deep Glimmse Network to capture temporal information. \cite{HCT} formulates MC-MOT as a compositional structure optimization problem, associating tracklets by appearance, geometry, and motion consistency. Starting with a scene node, a Spatial-Temporal Attributed Parse Graph is then constructed in \cite{STP}. The scene nodes are then decomposed into several tracklet nodes, containing different types of semantic attributes, such as appearance and action.

\vspace{-1em}
\paragraph{Graph-Based Methods}
Graph Neural Networks (GNN) \cite{GNN} and Graph Convolutional Networks (GCN) \cite{GCN} have also been extensively studied for MOT \cite{neuralsolver,TransMOT,polarmot,LMGP,DyGLIP} due to the flexibility of dynamical affinity association training. In \cite{polarmot}, it performs a standard graph model, passing the messages using localized polar feature representation. Nonetheless, it makes the association difficult when the spatial and temporal features are ambiguous and edges are all mixed up in one graph. \cite{TransMOT} first constructs a candidate graph, followed by Transformer \cite{Transformer} which serves as a feature encoder. However, they do not utilize spatial-temporal consistency in the graph. 
Combined with attention mechanism, the structural and temporal attention layers in \cite{DyGLIP} enable robust feature extraction for link prediction. It relies on single-camera tracker as input, leading to a sub-optimal solution in multi-views setting. The computation cost of dual attention layers is expensive. Graph Reconfiguration was not performed in \cite{DyGLIP}; instead, they aggregate objects in consecutive frames by directly adding new nodes and edges.
In \cite{LMGP}, multicut \cite{multicut1,multicut2,multicut3} is applied to MC-MOT to obtain the globally optimal solution. Likewise, their model relies on single-camera tracker to generate initial tracklets. Compared to our stage-wise optimization, \cite{LMGP} proposed a joint spatial-temporal optimization model. Mixing all spatial and temporal edges into one graph cannot fully leverage spatial and temporal consistency individually.

\section{Proposed Method}

Our method follows the tracking-by-detection paradigm \cite{tracking-by-detection}. In contrast to prior methods \cite{TRACTA,LMGP,DyGLIP,HCT}, our model does not rely on the off-the-shelf single-camera trackers trained by massive single-camera MOT datasets. Instead, the tracking and association problem is formulated into a link prediction problem on graph. The proposed ReST framework divides the MC-MOT process into two sub-tasks, Spatial Association and Temporal Association. Spatial Association focuses on matching objects across different views. The spatial graph concentrates on building spatial correlation between nodes. Temporal Association exploits temporal information, such as speed and time. The temporal graph extracts temporal features to associate objects across frames. These two modules take turns constructing spatial and temporal graphs for each frame. In addition, a novel Graph Reconfiguration module is proposed to reconfigure current spatial graph and temporal graph from previous timestamp as a new temporal graph. By this setting, the proposed graph model associates all detected objects frame by frame via extracting robust spatial and temporal-aware features. For the inference of each graph, we perform a post-processing module, which contains pruning and splitting, to fix association errors in time in each iteration. Compared with one single spatial-temporal graph model, two expert graph models can be trained separately, making it focus on extracting specific features and reduce the ambiguity of spatial-temporal correlations. The system framework of ReST is depicted in Figure~\ref{fig:framework}.

\subsection{Problem Formulation}\label{sub:3_1}

In MC-MOT, the goal is to track multiple objects across frames and views. Assume there are $C$ synchronous and static cameras that have overlapping fields of view (FoV). Define a graph model $G_t=(V_t,E_t)$ at time $t$, where $V_t$ is the vertex set and $E_t$ is the edge set.
Each node $v_i\in{V_t}$ represents one input detection and it contains the following information: camera ID $c_{v_i}\in{\mathbb{R}^1}$, timestamp $t_{v_i}\in{\mathbb{R}^1}$, object ID $o_{v_i}\in{\mathbb{R}^1}$, i.e. ground truth label, bounding-box position $b_{v_i}\in{\mathbb{R}^4}$, appearance feature $d_{v_i}\in{\mathbb{R}^{512}}$, geometry position $p_{v_i}\in{\mathbb{R}^2}$, and speed information $s_{v_i}\in{\mathbb{R}^2}$.
Let $I_{v_i}$ be the cropped image of $b_{v_i}$. The appearance feature $d_{v_i}$ can be obtained by 
\begin{equation}
    d_{v_i}=f_{ReID}(I_{v_i}),
\end{equation}
where $f_{ReID}$ is an off-the-shelf Re-Identification (ReID) model.
Denote $P_c$ as the projection function of camera $c$. It projects the foot point of the bounding-box from its camera view to a common ground plane (Section \ref{sub:projection} Appendix).
The geometry position $p_{v_i}$ and the speed information $s_{v_i}$ for the reference node $v_j$ can be calculated by 
\begin{equation}
    p_{v_i}=P_{c_{v_i}}(x+\frac{w}{2},y+h), \hspace{0.1in}
    s_{v_i}=\frac{p_{v_i}-p_{v_j}}{t_{v_i}-t_{v_j}},
\end{equation}
where $(x,y,w,h)$ represents the bounding box ${b_{v_i}}$, and $t_{v_i}>t_{v_j}\geq0$.

The relative distance of geometry position, appearance feature, and speed between any pair of nodes $v_i$ and $v_j$ can be defined by
\begin{equation*}
    \Delta{d_{ij}}=[\lVert d_{v_i}-d_{v_j}\rVert_1, 1-cosine\_similarity(d_{v_i},d_{v_j})],
\end{equation*}
\begin{equation*}
    \Delta{p_{ij}}=[\lVert p_{v_i}-p_{v_j}\rVert_1,\lVert p_{v_i}-p_{v_j}\rVert_2],
\end{equation*}
\begin{equation}\label{eq:dist}
    \Delta{s_{ij}}=[\lVert s_{v_i}-s_{v_j}\rVert_1,\lVert s_{v_i}-s_{v_j}\rVert_2].
\end{equation}
Following \cite{GNN-CCA}, $\Delta{p_{ij}}, \Delta{d_{ij}}, and \Delta{s_{ij}}$ are used as the initial edge features.

\subsection{Reconfigurable Spatial-Temporal Graph}\label{sub:3_2}

To better associate objects across views and frames, a novel reconfigurable graph framework is proposed for the inference stage. Our model follows the pipeline to achieve object tracking and association: perform Temporal Association to construct temporal graph at time $t$-1, Spatial Association to build spatial graph at time $t$ and then Graph Reconfiguration is applied to reconfigure a new temporal graph at time $t$.

\vspace{-1em}
\paragraph{Spatial Association}
Objects from different views at the current frame are first associated. In this stage, only spatial information is required to construct the spatial graph. The spatial graph is denoted as $G^S$. It concentrates on extracting spatial features for cross-view association. We denote all detected objects from camera $c$ at time $t$ as $B^t_c$. Given detected objects from all cameras at time $t$, the vertex set $V^S_t$ of spatial graph $G^S_t$ can be defined by
\begin{equation}\label{eq:init_node_SG}
    V^S_t=\bigcup_{i=1}^{C}B^t_i.
\end{equation}
The vertex set $V^S_t$ is composed of all detected objects across cameras at time $t$. We denote the adjacency matrix of $G^S_t$ as $A^S_t=[a^S_{ij}]$. The initial edge construction is defined by
\begin{equation}\label{eq:init_edge_SG}
    a^S_{ij}=
      \begin{cases}
      1, & \text{if $c_{v_i}\neq{c_{v_j}}$} \\
      0, & \text{otherwise}
      \end{cases}.
\end{equation}
That is, there is an edge between node $v_i$ and $v_j$ if both nodes are from different cameras. Once the spatial graph is constructed, the initial node feature $h^0_{v_i}$ and initial edge feature $h^0_{e_{ij}}$ can be defined by
\begin{equation*}
    h^0_{v_i}=f^v_{FE}(d_{v_i}),
\end{equation*}
\begin{equation}\label{eq:init_feature_SG}
    h^0_{e_{ij}}=f^e_{FE}([\Delta{p_{ij}},\Delta{d_{ij}}]),
\end{equation}
where $f^v_{FE}$ is a node feature encoder and $f^e_{FE}$ is an edge feature encoder. Both are implemented by Multi-Layer Perceptron (MLP). The operator $[\cdot , \cdot]$ denotes the concatenation of two terms.
Thus, $h^0_{v_i}$ and $h^0_{e_{ij}}$ are used as input of Message Passing Network (MPN) to extract edge features. When final edge features are extracted, link prediction is performed to construct the final spatial graph as the result of object association. Followed by post-processing module, the spatial graph can be further refined. Line 3 to line 9 in Algorithm~\ref{alg:inference} present the steps of spatial association. The details of MPN and link prediction will be described in subsection \ref{sub:MPN} and \ref{sub:link prediction}.

\vspace{-1em}
\paragraph{Temporal Association}
In this stage, objects are associated from different frames by time, without using any camera information. In other words, the temporal graph is view-invariant and cares more about temporal correlation. Given temporal graph $G^T_{t}$ at time $t$, the initial node and edge features can be computed by
\begin{equation*}
    h^0_{v_i}=f^v_{FE}([d_{v_i},p_{v_i}]),
\end{equation*}
\begin{equation}\label{eq:init_feature_TG}
    h^0_{e_{ij}}=f^e_{FE}([\Delta{p_{ij}},\Delta{d_{ij}},\Delta{s_{ij}}]).
\end{equation}
Note that we append an extra speed term in the input of edge feature to capture relative motion and direction between nodes. Following similar steps in $G^S_t$, MPN, link prediction, and post-processing module are performed, as indicated from line 12 to 17 in Algorithm \ref{alg:inference}.
For temporal graph $G^T_t$, tracklet ID is assigned to each node within the same connected component as the tracking results at time $t$.

\vspace{-1em}
\paragraph{Graph Reconfiguration}
After the association stage, several connected components representing the same object can be obtained. In order to bridge two association stages, our model reconfigures two graphs into a new Temporal Graph. $G^T_t$ is denoted as the temporal graph at time $t$. $G^T_t$ is reconfigured from spatial graph $G^S_t$ and temporal graph $G^T_{t-1}$. From $G^S_t$ and $G^T_{t-1}$, all nodes in the same connected components are aggregated into one node. After the aggregation, a new vertex set is formed for $G^T_t$.
Let $H(G)=\{H_1,...,H_n\}$ be the set of connected components in $G$.
All node information within one connected component is averaged and serves as the initial features, i.e.  
\begin{equation*}
    d_v=\frac{\sum_{v\in{H_i}}d_{v}}{|H_i|},
\end{equation*}
\begin{equation}
    p_v=\frac{\sum_{v\in{H_i}}p_{v}}{|H_i|}.
\end{equation}
where $H_i \in H(G^S_t)\cup H(G^T_{t-1})$.
Once the vertex set of $G^T_t$ is determined, the edge of $G^T_t$ can be defined by
\begin{equation}\label{eq:init_edge_TG}
    a^T_{ij}=
      \begin{cases}
      1, & \text{if $t_{v_i}\neq{t_{v_j}}$} \\
      0, & \text{otherwise}
      \end{cases},
\end{equation}
where $a^T_{ij}\in{A^T_t}$ is adjacency matrix of $G^T_t$. The edges exist in $G^T_t$ if two nodes are from different time frames. The complete steps in the inference are given in Algorithm~\ref{alg:inference}. 


\begin{algorithm}[h]
  \caption{Inference Algorithm}\label{alg:inference}
  \begin{algorithmic}[1]
    \Statex \textbf{Input:} temporal graph $G^T_{t-1}$, detection set $B_c^t$ from all $C$ views at time $t$.
    \Statex \textbf{Output:} temporal graph $G^T_t$, tracking result at time $t$.
    \State construct spatial graph $G^S_t$ through $B^t_c$
    \State compute initial feature $h^0_{v_i},h^0_{e_{ij}}$ for $G^S_t$
    \For{$l=1$ to $L$}
      \State $h^l_{v_i},h^l_{e_{ij}}=\text{MPN}(G^S_t,h^{l-1}_{v_i},h^{l-1}_{e_{ij}})$
    \EndFor
    \State $\hat{y}^L_{e_{ij}}=f_{CLS}(h^L_{e_{ij}})$
    \State $G^S_t = \text{post-processing}(G^S_t,\hat{y}^L_{e_{ij}})$
    \If{$t>0$}
      \State $G^T_t=\text{reconfiguration}(G^S_t,G^T_{t-1})$
      \State compute initial feature $h^0_{v_i},h^0_{e_{ij}}$ for $G^T_t$
      \For{$l=1$ to $L$}
        \State $h^l_{v_i},h^l_{e_{ij}}=\text{MPN}(G^T_t,h^{l-1}_{v_i},h^{l-1}_{e_{ij}})$
      \EndFor
      \State $\hat{y}^L_{e_{ij}}=f_{CLS}(h^L_{e_{ij}})$
      \State $G^T_{t} = \text{post-processing}(G^T_t, \hat{y}^L_{e_{ij}})$
    \EndIf
  \end{algorithmic}
\end{algorithm}

\vspace{-1em}
\begin{algorithm}[h]
  \caption{Post-Processing Algorithm}
  \label{alg:pp}
  \begin{algorithmic}[1]
    \Statex \textbf{Input:} $G_t$: graph at time $t$, $\hat{y}^L_{e_{ij}}$: confidence score.
    \Statex \textbf{Output:} $G_t$: refined graph at time $t$.
    \State $G_t=\text{pruning}(G_t, \hat{y}^L_{e_{ij}})$
    \State $G_t=\text{splitting}(G_t)$
    \If{$G_t$ is temporal graph}
      \State assigning tracklet ID
    \EndIf
  \end{algorithmic}
\end{algorithm}

\vspace{-1em}
\subsection{Post-Processing}\label{sub:3_3}

In post-processing, the objective is to refine the graph output. Since the vertices in the same connected component represent objects of the same ID, the connected component may contain vertices with different object IDs. Additional constraints can be included to reduce incorrect ID assignments. The post-processing is divided into three steps: pruning, splitting, and assigning object ID.

\vspace{-1em}
\paragraph{Pruning}
Confidence score of each edge predicted by our model is used to prune the graph. If the score is greater than a given threshold $\varepsilon$, the edge is kept. Otherwise, it is removed. After pruning, edges with weak confidence are removed to improve the correctness of connected components.

\vspace{-1em}
\paragraph{Splitting} 
Similar to \cite{GNN-CCA}, a few physical constraints and assumptions can be employed to further optimize each connected component. In spatial graph, assume that an object can only appear in each camera once, there are at most $C$ nodes in each connected component and each node can be connected to at most $C-1$ nodes. Therefore, the constraints of spatial graph can be defined as
\begin{equation}\label{eq:constraint1}
     |V(H^S_i)|\leq{C},
\end{equation}
where ${H^S_i}\in{H(G^S)}$. For every node ${v}\in{H^S_i}$, we have
\begin{equation}\label{eq:constraint2}
    \text{degree}(v)\leq{C-1}.
\end{equation}
Each node in temporal graph can be connected to at most $M-1$ nodes, where $M$ is the temporal window size.
\begin{equation}\label{eq:constraint3}
    \forall{v}\in{H^T_i}\in{H(G^T)},\text{degree}(v)\leq{M-1}.
\end{equation}
For any connected component violating Eq.\eqref{eq:constraint1}-\eqref{eq:constraint3}, the edge with the lowest confidence score is removed. The operation is performed recursively until all constraints are satisfied.

\vspace{-1em}
\paragraph{Assigning tracklet ID}
After the post-processing for temporal graph is finished, tracklet IDs are assigned to nodes in the current frame. In practice, one node inherits the ID from nodes that are already in the same connected component, otherwise it is assigned a new ID (Section \ref{sub:trackletID} Appendix). The post-processing steps are given in Algorithm~\ref{alg:pp}.

\subsection{Model Training}\label{sub:3_4}

\begin{figure}[t]
\begin{center}
\includegraphics[width=1\linewidth]{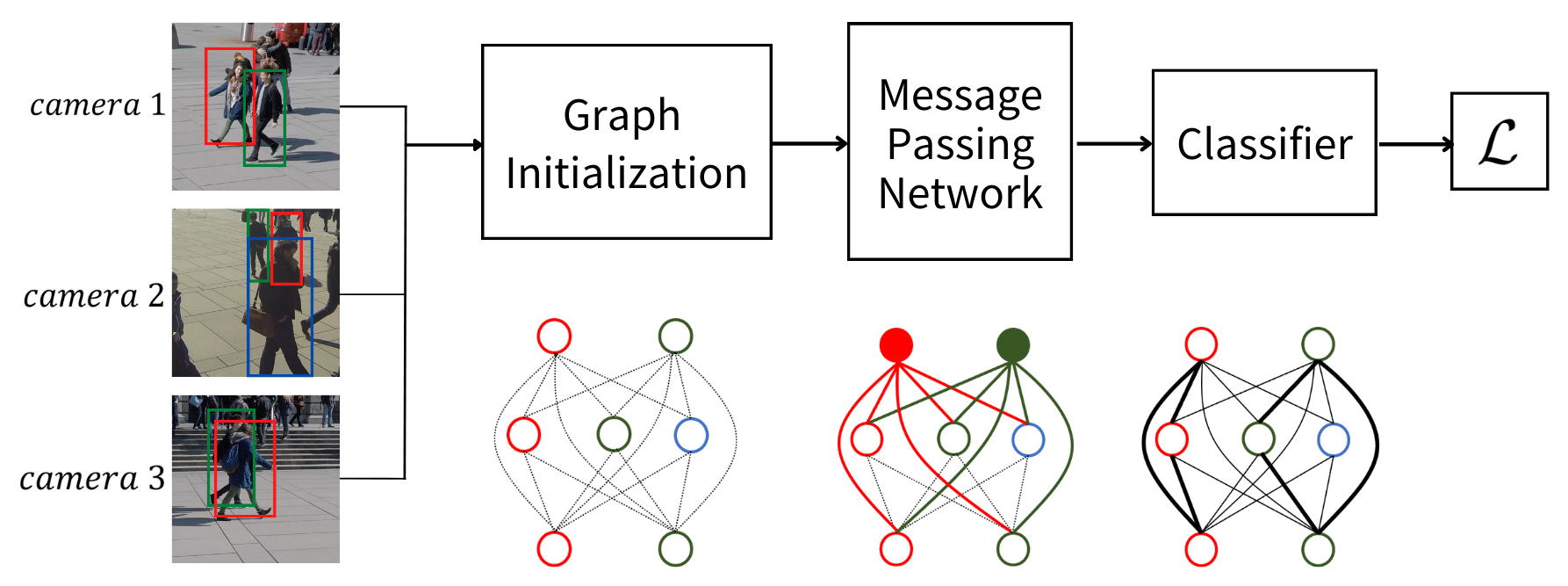}
\end{center}
   \caption{Graph model training. Given different input detection and constraints, we construct and train $G^S$ and $G^T$, respectively. This figure shows an example of input detection for spatial graph.}
\vspace{-1em}
\label{fig:graph_training}
\end{figure}

In this subsection, the details of Message Passing Network, link prediction, and training scheme are described, as depicted in Figure~\ref{fig:graph_training}.

\vspace{-1em}
\subsubsection{Message Passing Network}\label{sub:MPN}
Given the initial feature $h^0_{v_i}$ and $h^0_{e_{ij}}$ of a graph, we follow the standard framework of MPN \cite{neuralsolver,polarmot,GNN-CCA} and perform a fixed number of graph updates to obtain enhanced feature representation. Specifically, there are two steps in a graph update iteration; namely, Edge Update and Node Update.

\paragraph{Edge Update}
At each message passing iteration $l=1,...,L$, edge feature is firstly updated by aggregating its source node feature and destination node feature as:
\begin{equation}\label{eq:edge_update}
    h^l_{e_{ij}}=f^e_{ME}([h^{l-1}_{v_i},h^{l-1}_{e_{ij}},h^{l-1}_{v_j}]),
\end{equation}
where $f^e_{ME}$ is an edge message encoder. MLP is exploited to encode the original message into a high-dimensional feature space.

\vspace{-1em}
\paragraph{Node Update}
After updating the edge feature, we then update each node by the messages sent from its neighbor nodes as:
\begin{equation}\label{eq:node_update}
    h^l_{v_i}=\sum_{j\in{N(v_i)}}m^l_{ij},
\end{equation}
where $N(v_i)$ denotes the neighbor nodes of $v_i$, and the message term can be computed by
\begin{equation}
    m^l_{ij}=f^v_{ME}([h^{l-1}_{v_j},h^l_{e_{ij}}]),
\end{equation}
where $f^v_{ME}$ is a node message encoder similar to $f^e_{ME}$.

\vspace{-1em}
\subsubsection{Link Prediction}
\label{sub:link prediction}

After MPN, enhanced edge features can be obtained for link prediction. It aims to decide whether an edge should be kept or removed in a graph. Specifically, a binary classifier is cascaded to MPN. Given the edge feature from iteration $l$, the classifier outputs a confidence score:
\begin{equation}
    \hat{y}^l_{e_{ij}}=f_{CLS}(h^l_{e_{ij}}),
\end{equation}
where $f_{CLS}$ is a binary classifier implemented by MLP followed by a softmax layer. In the inference stage, the confidence score $\hat{y}^L_{e_{ij}}$ at the last iteration is used for pruning.

\vspace{-1em}
\subsubsection{Training Scheme}
In model training, the spatial graph and temporal graph are trained independently to learn spatial and temporal-aware feature representation. To train the spatial graph, training input is all detections from different views at the same time frame. For temporal graph, training input only contains detections from different frames of the same camera $c$ as
\begin{equation}\label{eq:init_node_TG}
    V^T_t=\bigcup_{i=0}^{M-1}B_c^{t-i}.
\end{equation}

For both graphs, the ReID model $f_{ReID}$ is frozen during the training process. $f^v_{FE}, f^e_{FE}, f^v_{ME}, f^e_{ME}$, and $f_{CLS}$ are trainable MLPs (Section \ref{sub:network} Appendix). Focal Loss \cite{focalloss} is exploited to calculate the loss between ground-truth label and predicted label at each message passing iteration $l$, given by
\begin{equation}\label{eq:loss}
    \mathcal{L}=\sum^L_{l=1}\sum_{e_{ij}\in{E^S\cup{E^T}}}FL(\hat{y}^l_{e_{ij}},y_{e_{ij}}),
\end{equation}
where $y_{e_{ij}}$ is ground truth label and its value equals 1 if $v_i$ and $v_j$ have the same object ID, i.e. $o_{v_i}=o_{v_j}$. Otherwise, it is 0.

In this way, our graph model can effectively learn how to associate two nodes spatially and temporally. Compared with other single spatial-temporal graph methods \cite{neuralsolver,polarmot,LMGP,DyGLIP}, our graph can focus on learning more discriminating spatial and temporal features to cope with challenging multi-object tracking scenarios.

\section{Experimental Results}

In this section, we demonstrate our model performance on several benchmark datasets. Detailed implementation settings and ablation studies are presented. For evaluation of MC-MOT methods, ID score \cite{IDF1}, i.e. IDF1, and the standard CLEAR MOT metrics \cite{CLEARMOT}, including MOTA, MOTP, Mostly Tracked (MT), and Mostly Lost (ML), are employed for a fair comparison. Experimental comparisons with the state-of-the-art MC-MOT methods are also presented in this section.

\subsection{Datasets}

Our experiments are conducted on three multi-view multi-object tracking datasets under diverse environmental conditions, such as illumination, density, and detection quality. All video sequences have synchronous and calibrated cameras with a certain ratio of overlapping FoV.

\begin{figure*}
\begin{center}
\begin{subfigure}[b]{0.28\linewidth}
         \centering
         \includegraphics[width=1\linewidth]{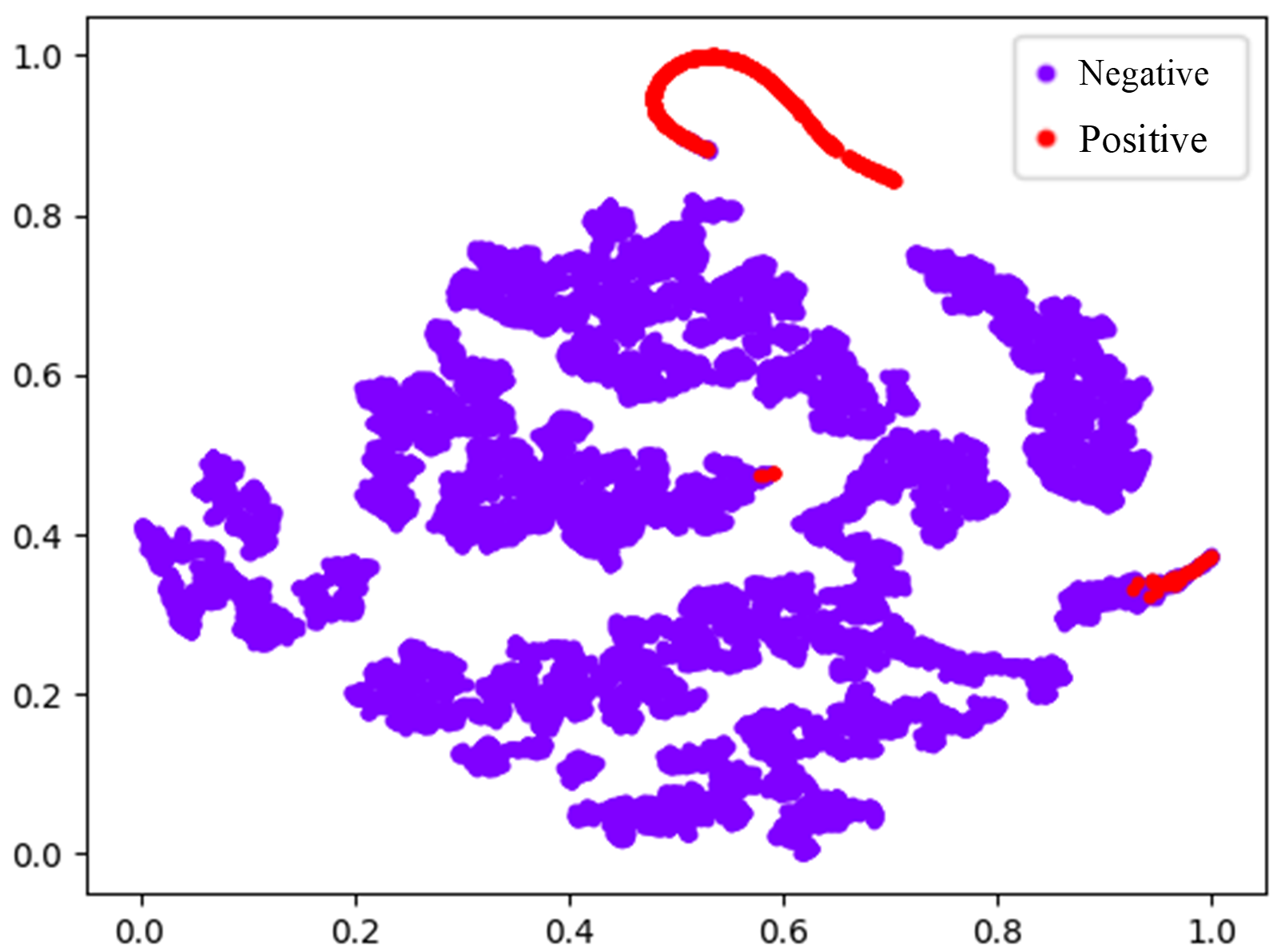}
         \caption{Single Graph}
         \label{fig:edge-a}
\end{subfigure}
\hfill
\begin{subfigure}[b]{0.28\linewidth}
         \centering
         \includegraphics[width=1\linewidth]{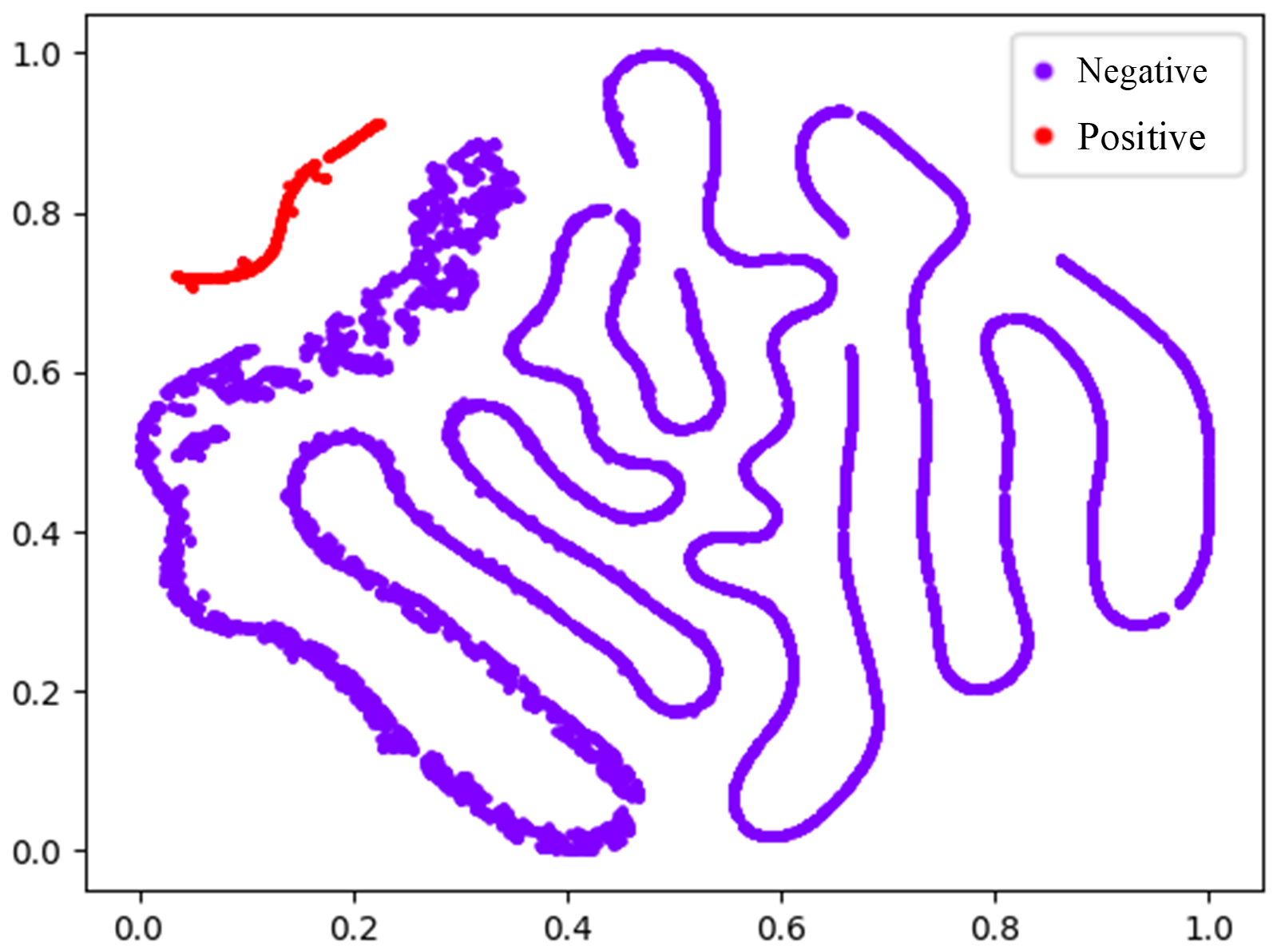}
         \caption{Spatial Graph}
         \label{fig:edge-b}
\end{subfigure}
\hfill
\begin{subfigure}[b]{0.28\linewidth}
         \centering
         \includegraphics[width=1\linewidth]{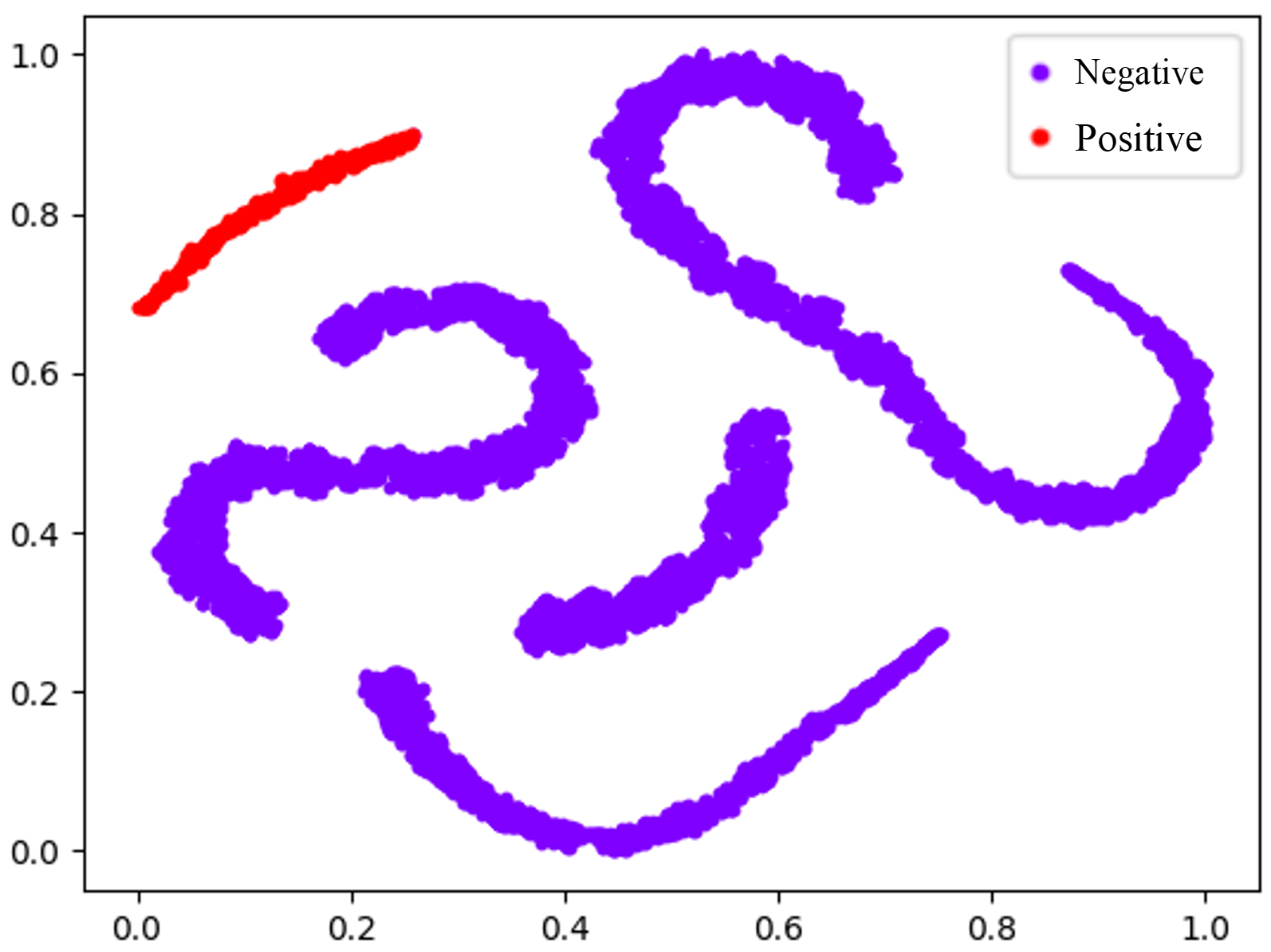}
         \caption{Temporal Graph}
         \label{fig:edge-c}
\end{subfigure}
\end{center}
   \caption{Edge feature clustered by t-SNE\cite{tsne}. (a) Edge feature of single graph model. (b) Edge feature of $G^S$. (c) Edge feature of $G^T$. We sample 20 graphs and plot all edge features on figures. Positive and negative are ground truth labels determined by whether the edge comes from the same object. There is a more clear boundary in both spatial graph and temporal graph than single graph, leading to more accurate classification results.}
\vspace{-1.5em}
\label{fig:edge}
\end{figure*}

\vspace{-1em}
\paragraph{Wildtrack \cite{wildtrack}}
It is considered the most challenging dataset with 7 cameras, having the most occlusion problems and the highest density. Specifically, there are about 25 people standing and walking around at each frame on average. We follow the common setting as \cite{wildtrack,Engilberge_2023_WACV,mvdetr,mvdet}, which is trained on the first 360 frames and tested on the last 40 frames.

\vspace{-1em}
\paragraph{CAMPUS \cite{HCT}}
All of the sequences in CAMPUS are reported in our results. People are doing all kinds of sports in \textit{Garden 1}, which means the capability to capture diverse motion is crucial. \textit{Garden 2} is a relatively sparse sequence. People are often occluded by cars in the \textit{Parkinglot} sequence, making it hard to recover from different views. \textit{Auditorium} is a sequence recorded in two scenes that we use to validate our model's ability to apply in a non-overlapping FoV scenario.

\vspace{-1em}
\paragraph{PETS-09 \cite{PETS09}}
The results of S2.L1 sequence are reported for a complete comparison with other methods. Similar to CAMPUS, there are less than 10 people at each frame on average. Although it is not as dense as Wildtrack, the low video quality, e.g. various illumination and cameras far away from people, is the most challenging part.

\subsection{Implementation Details}

OSNet \cite{OSNet} is exploited as ReID model to extract appearance feature, which outputs a 512-D feature vector. Input image for OSNet is object bounding-boxes cropped from the original frame and resized to 256$\times$128. In graph model, the dimension of node feature is 32-D, while edge feature is 6-D. We run $L=4$ message passing iterations in all experiments, and then output edge feature for link prediction classifier, which is also a 6-D feature vector.

For model training, we use ground truth detection as input and set the temporal window size $M$=3 for $G^T$. The model is trained by the following settings: Adam optimizer \cite{Adam} is employed to run 100 epochs. Warm-up learning rate is set starting from 0 to 0.01 in the first 10 epochs.
We randomly drop detection for data augmentation to mimic false negative cases. In the inference stage, we use detection from MVDeTr \cite{mvdetr} in Wildtrack and provided detection in the other datasets for a fair comparison. The weights of graph model with the highest validation performance are used for testing. The pruning threshold $\varepsilon$ is set to 0.9 to retain high confident edges only.

\subsection{Results of MC-MOT}

\begin{table}
  \begin{center}
  \resizebox{\columnwidth}{!}{%
  \begin{tabular}{c|c|c|c|c|c}
    \hline
    Method & IDF1$\uparrow$ & MOTA$\uparrow$ & MOTP$\uparrow$ & MT$\uparrow$ & ML$\downarrow$ \\
    \hline\hline
    KSP-DO \cite{wildtrack}           & 73.2 & 69.6 & 61.5 & 28.7 & 25.1 \\
    KSP-DO-ptrack \cite{wildtrack}    & 78.4 & 72.2 & 60.3 & 42.1 & 14.6 \\
    GLMB-YOLOv3 \cite{GLMB}      & 74.3 & 69.7 & 73.2 & 79.5 & 21.6 \\
    GLMB-DO \cite{GLMB}          & 72.5 & 70.1 & 63.1 & \textbf{93.6} & 22.8 \\
    T-Glimpse \cite{tglimmpse}        & 77.8 & 72.8 & 79.1 & 61.0 & 4.9 \\
    T-Glimpse Stack \cite{tglimmpse}  & 81.9 & 74.6 & 78.9 & 65.9 & 4.9 \\
    \hline
    Ours                & \textbf{85.7} & \textbf{81.6} & \textbf{81.8} & 79.4 & \textbf{4.7}\\
    \hline
  \end{tabular}%
  }
  \end{center}
  \caption{Evaluation results on Wildtrack. We achieve state-of-the-art performance with 3.8\% and 7.0\% progress on IDF1 and MOTA.}
  \label{tb:Wildtrack}
  \vspace{-1.5em}
\end{table}
\begin{table}
  \begin{center}
  \resizebox{\columnwidth}{!}{%
  \begin{tabular}{c|c|c|c|c|c}
    \hline
    Sequence & Method & MOTA$\uparrow$ & MOTP$\uparrow$ & MT$\uparrow$ & ML$\downarrow$ \\
    \hline\hline
    \multirow{6}{*}{Garden 1}
      & HCT \cite{HCT}      & 49   & 71.9 & 31.3 & 6.3 \\
      & STP \cite{STP}      & 57   & 75   & -    & -   \\
      & TRACTA \cite{TRACTA}   & 58.5 & 74.3 & 30.6 & 1.6 \\
      & DyGLIP \cite{DyGLIP}   & 71.2 & 91.6 & 31.3 & \textbf{0.0} \\
      & LMGP \cite{LMGP}     & 76.9 & 95.9 & 62.9 & 1.6 \\
      \cline{2-6}
      & Ours        & \textbf{77.6} & \textbf{99.1} & \textbf{100.0} & \textbf{0.0} \\
    \hline
    \multirow{6}{*}{Garden 2}
      & HCT \cite{HCT}      & 25.8 & 71.6 & 33.3 & 11.1 \\
      & STP \cite{STP}      & 30   & 75   & -    & -    \\
      & TRACTA \cite{TRACTA}   & 35.5 & 75.3 & 16.9 & 11.3 \\
      & DyGLIP \cite{DyGLIP}   & \textbf{87.0} & 98.4 & 66.7 & \textbf{0.0} \\
      \cline{2-6}
      & Ours        & 86.0 & \textbf{99.9} & \textbf{100.0} & \textbf{0.0} \\
    \hline
    \multirow{6}{*}{Parkinglot}
      & HCT \cite{HCT}      & 24.1 & 66.2 &  6.7 & 26.6 \\
      & STP \cite{STP}      & 28   & 68   & -    & -    \\
      & TRACTA \cite{TRACTA}   & 39.4 & 74.9 & 15.5 & 10.3 \\
      & DyGLIP \cite{DyGLIP}   & 72.8 & 98.6 & 26.7 & \textbf{0.0}  \\
      & LMGP \cite{LMGP}     & \textbf{78.1} & 97.3 & 62.1 & \textbf{0.0}  \\
      \cline{2-6}
      & Ours        & 77.7 & \textbf{99.8} & \textbf{100.0} & \textbf{0.0} \\
    \hline
    \multirow{5}{*}{Auditorium}
      & HCT \cite{HCT}      & 20.6 & 69.2 &  33.3 & 11.1 \\
      & STP \cite{STP}      & 24   & 72   & -    & -    \\
      & TRACTA \cite{TRACTA}   & 33.7 & 73.1 & 37.3 & 20.9 \\
      & DyGLIP \cite{DyGLIP}   & \textbf{96.7} & \textbf{99.5} & \textbf{95.2} & \textbf{0.0}  \\
      \cline{2-6}
      & Ours        & 81.2 & 98.8 & 92.1 & \textbf{0.0} \\
    \hline

  \end{tabular}%
  }
  \end{center}
  \caption{Evaluation results on CAMPUS. Our model perfectly tracked all people most of time (Figure \ref{fig:fig1-b}), leading to perfect scores on MT and ML and competitive results on MOTA.}
  \vspace{-1.5em}
  \label{tb:Campus}
\end{table}
\begin{table}
  \begin{center}
  \resizebox{\columnwidth}{!}{%
  \begin{tabular}{c|c|c|c|c|c}
    \hline
    Method & Online & MOTA$\uparrow$ & MOTP$\uparrow$ & MT$\uparrow$ & ML$\downarrow$ \\
    \hline\hline
    KSP  \cite{KSP}    &  & 80 & 57 & - & -\\
    TRACTA \cite{TRACTA}  &  & 87.5 & 79.2 & - & -\\
    DyGLIP \cite{DyGLIP} & \checkmark  & 93.5 & 94.7 & - & -\\
    STVH \cite{STVH}    & & 95.1 & 79.8 & \textbf{100.0} & \textbf{0.0} \\
    MLMRF \cite{MLMRF}  & & 96.8 & 79.9 & \textbf{100.0} & \textbf{0.0} \\
    LMGP \cite{LMGP}    & & \textbf{97.8} & 82.4 & \textbf{100.0} & \textbf{0.0} \\
    \hline
    Ours       & \checkmark & 92.3 & \textbf{99.7} & \textbf{100.0} & \textbf{0.0} \\
    \hline
  \end{tabular}%
  }
  \end{center}
  \caption{Evaluation results on PETS-09 sequence S2.L1. Check mark indicates the online method. When compared with other offline methods, our method still achieves very competitive performance.}
  \label{tb:PETS09}
  \vspace{-1em}
\end{table}

We report our model performance on Wildtrack, CAMPUS, and PETS-09, in Table \ref{tb:Wildtrack}, Table \ref{tb:Campus}, and Table \ref{tb:PETS09}, respectively. On Wildtrack, we compare with other online approaches using detector as input. Our results achieve state-of-the-art performance with 3.8\% and 7.0\% higher than the second place on IDF1 and MOTA. 
Our experimental results on CAMPUS outperform other approaches on most metrics. One can notice that, for MT and ML, our results on overlapping FoV sequences achieve 100 and 0, respectively. Even in the sequence with non-overlapping FoV, our model still performs well. This is because our method properly leverages spatial and temporal consistency, which allows steady tracking on each object. The results indicate that our method is suitable for handling fragmented tracklet due to occlusion.
On PETS-09, our method is competitive compared with other offline methods.

\subsection{Ablation Study}
To validate the robustness of our model, we conduct several ablation studies in this section.

\vspace{-1em}
\paragraph{Design of Input Feature}
Table \ref{tb:input_feature} presents the impact of different input features. If appearance, projection, or speed feature is removed for both nodes and edges, the performance drop goes up. In this study, the projection term is crucial to our method. As for the speed term, it is important for Temporal Association since it provides motion information. There is only a small drop in performance if the appearance feature is removed. Although ReID feature helps to associate object with its appearance, our model does not heavily rely on it. Our method presents better generalization when the illumination or appearance changes drastically across datasets.

\begin{table}
  \begin{center}
  \resizebox{\columnwidth}{!}{%
  \begin{tabular}{c|c|c|c|c}
    \hline
    Appearance & Projection & Speed & IDF1$\uparrow$ & MOTA$\uparrow$ \\
    \hline\hline
     \checkmark &  & \checkmark & 61.5 & 77.2 \\
    \checkmark & \checkmark &  & 86.9 & 94.8 \\
     & \checkmark & \checkmark & 89.2 & 95.0 \\
    \hline
    \checkmark & \checkmark & \checkmark & \textbf{91.6} & \textbf{97.0} \\
    \hline
  \end{tabular}%
  }
  \end{center}
  \caption{Tracking performance between different combination of input feature on Wildtrack. We use ground-truth detection to focus on the impact on association stage.}
  \label{tb:input_feature}
  \vspace{-1.5em}
\end{table}

\vspace{-1em}
\paragraph{Separated vs. Unified Graph Models}
To extract spatial and temporal-aware feature for better association, spatial graph and temporal graph are trained separately with different training input. To compare with the feature learning by one unified spatial-temporal graph, a unified spatial-temporal graph model is trained without any edge constraints and the input data is chunks of frames containing detection across views and frames. Edge features with positive and negative labels are visualized by applying t-SNE \cite{tsne}. As depicted in Figure \ref{fig:edge}(a), the negative edge features are messed with positive edge features in some areas, leading to more failure cases in MC-MOT. In Figure~\ref{fig:edge}(b)(c), a clear boundary between positive and negative edges can be observed. That is, with separate spatial and temporal graphs and their own constraints, the learned edge features are more effective for data association.

\begin{figure}[t]
\begin{center}
\includegraphics[width=0.93\linewidth]{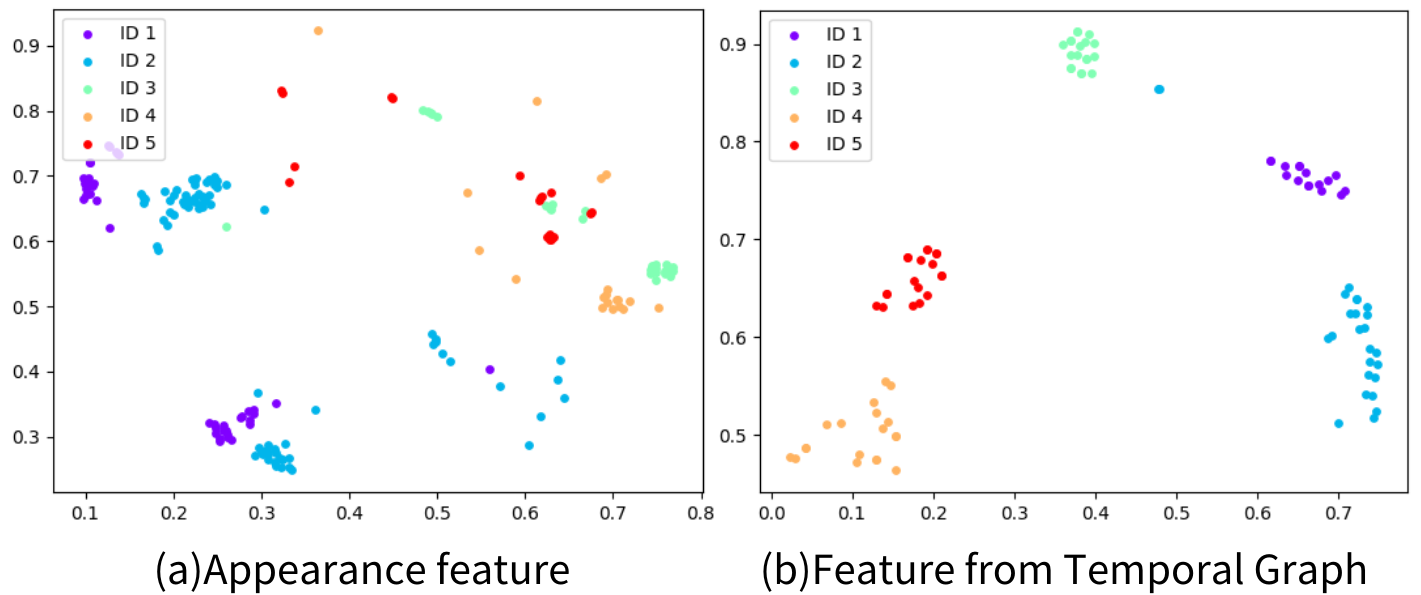}
\end{center}
   \caption{Node feature clustered by t-SNE. (a) Appearance feature extracted by \cite{OSNet}. (b) Feature of ReST model. Each node represents one detection at certain frame. We sample 5 people from 20 consecutive frames in both figures.}
\label{fig:tSNE}
\vspace{-1em}
\end{figure}

\vspace{-1em}
\paragraph{Robust Feature Representation}
The quality of feature representation significantly affects the association accuracy in MC-MOT. Precise tracklets can be predicted given better feature representation. In Figure \ref{fig:tSNE}, node feature embedding is visualized via t-SNE. In contrast to features extracted by ReID model \cite{OSNet}, features learned by ReST have better between-class separation and within-class aggregation. Therefore, we can better discriminate and associate objects by using ReST.

\vspace{-1em}
\paragraph{Cross-dataset Testing}
We conduct cross-dataset testing to validate the generalization ability of our model. Specifically, we load model weights trained on Wildtrack and perform inference on CAMPUS dataset. As shown in Table \ref{tb:general}, there is a significant improvement in MOTA up to 15.5\%. Although Wildtrack has fewer frames than others, it supplies more abundant information for model training, e.g. more difficult occlusion cases and more diverse motion patterns. In addition, lower frame rate makes it extract non-static speed information between frames than other scenes. Furthermore, our model does not heavily rely on appearance feature as shown in Table \ref{tb:input_feature}, leading to better generalization results.

\begin{table}
  \begin{center}
  \resizebox{\columnwidth}{!}{%
  \begin{tabular}{c|c|c|c}
    \hline
    Testing Sequence & Training Sequence & MOTA$\uparrow$ & MOTP$\uparrow$ \\
    \hline\hline
    \multirow{2}{*}{Garden 1}
      & Garden 1     & 77.6 & \textbf{99.1} \\
      & Wildtrack    & \textbf{93.1} & 90.9\\
    \hline
    \multirow{2}{*}{Garden 2}
      & Garden 2     & 86.0 & 99.9\\
      & Wildtrack    & \textbf{90.2} & \textbf{100.0}\\
    \hline
    \multirow{2}{*}{Parkinglot}
      & Parkinglot  & 77.7 & \textbf{99.8}\\
      & Wildtrack  & \textbf{92.4} & \textbf{99.8}\\
    \hline
    \multirow{2}{*}{Auditorium}
      & Auditorium  & 81.2 & 98.8\\
      & Wildtrack  & \textbf{95.8} & \textbf{98.9}\\
    \hline
  \end{tabular}%
  }
  \end{center}
  \caption{Cross-dataset testing on CAMPUS. We train the model on Wildtrack and test each sequence in CAMPUS.}
  \label{tb:general}
  \vspace{-1.5em}
\end{table}

\section{Conclusion}
In this paper, we propose a novel reconfigurable graph model for MC-MOT. A two-stage association scheme is proposed via Spatial Association and Temporal Association. It first associates objects across different views at the same frame using spatial graph. Followed by Graph Reconfiguration module which aggregates the nodes within the same connected component to simplify the graph and reconfigures it into a new temporal graph. Lastly, Temporal Association is applied to match objects across frames to accomplish online tracking. The spatial graph and temporal graph are independently trained to concentrate on spatial and temporal-domain feature learning, respectively.
As shown in the experimental results, we can learn more discriminating features for object association, leading to state-of-the-art performance on Wildtrack and competitive results on other datasets compared with other offline methods. In the future, we plan to investigate more flexible graph reconfiguration of spatial/temporal or spatial-temporal graph models for MC-MOT.
\section{Acknowledgements}
This work was supported in part by the National Science and Technology Council, Taiwan under grants NSTC-111-2221-E-007-106-MY3, NSTC-111-2634-F-007-010, and NSTC-111-2634-F-194-003-. We also thank National Center for High-performance Computing in Taiwan for providing computational and storage resources.


{\small
\bibliographystyle{ieee_fullname}
\bibliography{egpaper_final}
}

\newpage
\appendix
\renewcommand\thesection{\Alph{section}}
\renewcommand\thesubsection{\thesection.\arabic{subsection}}
\renewcommand\thesubsubsection{\thesubsection.\arabic{subsubsection}}

\section*{Supplementary Material}
In this supplementary material, we further describe additional details to complement our proposed reconfigurable graph model.
Firstly, the calculation of the projection function is detailed in Section \ref{sub:projection}, followed by tracklet ID assignment steps in Section \ref{sub:trackletID}. 
Afterward, our network architecture and model complexity are described in Section \ref{sub:network}.
Section \ref{sub:pp} shows the effectiveness of post-processing module.
Additionally, we provide visualization of our graph model to better realize two association stages in Section \ref{sub:graph-vis}. 
Qualitative results are shown in Section \ref{sub:qualitative} to explain how we fix the fragmented tracklet problems. 
Lastly, we demonstrate the proposed ReST tracker in the attached link.
\section{Calculation of Geometry Position}\label{sub:projection}
The geometry position of node $v_i$ can be obtained by projecting its estimated foot point from the camera view to a common ground plane via a projection function. The projection function $P_{c_{v_i}}$ is based on camera calibration parameters and derived from perspective projection:
\begin{equation}
    x_{img}=K[R\quad t]x_{world}=Px_{world},
\end{equation}
where $x_{img}$ and $x_{world}$ denote the positions in 2D image and 3D world represented in homogeneous coordinates, respectively, and $K$ is the intrinsic matrix, with rotation matrix $R$ and translation vector $t$ determined by the extrinsic parameters. Assuming $z=0$ as the common ground plane, the $3\times4$ projection matrix $P$ is reduced to be a $3\times3$ homography matrix $H$. Therefore, the position $p_{v_i}$ on the ground plane can be calculated by
\begin{equation}
    p_{v_i}=H^{-1}x_{img},
\end{equation}
where $x_{img}$ is the person's foot point, estimated by the position, width, and height of the bounding box.

\section{Tracklet ID Assignment}\label{sub:trackletID}
\begin{figure}[t]
\begin{center}
\begin{subfigure}[b]{0.51\linewidth}
         \centering
         \includegraphics[width=1\linewidth]{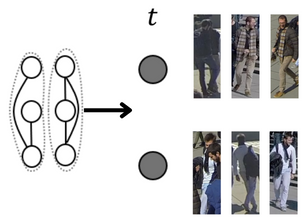}
         \caption{}
         \label{fig:assign-a}
\end{subfigure}
\hfill
\begin{subfigure}[b]{0.44\linewidth}
         \centering
         \includegraphics[width=1\linewidth]{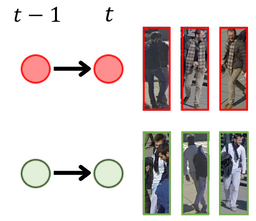}
         \caption{}
         \label{fig:assign-b}
\end{subfigure}
\end{center}
   \caption{Tracklet ID Assignment. (a) In Graph Reconfiguration, we aggregate a list of bounding-boxes and camera IDs from its original connected component into a node (black solid node). (b) After Temporal Association, each node at time $t$ and its associated detections will inherit ID from the previous node or be assigned to a new ID.}
\label{fig:assign}
\vspace{-0.5em}
\end{figure}

In this section, we explain the details of tracklet ID assignment steps.
In the spatial graph, one node $v_i$ represents one detection, including bounding-box location $b_{v_i}$ and camera ID $c_{v_i}$.
In Graph Reconfiguration stage, we save all bounding-boxes and their respective camera IDs within the same connected component of spatial graph $G^S_t$, and then aggregate them into a node of $G^T_t$ (Figure \ref{fig:assign-a}).
The last step of post-processing in Temporal Association is assigning tracklet ID.
As shown in Figure \ref{fig:assign-b}, if an aggregated node at the current frame is connected to a previous node, all of its detection, i.e. bounding-box and camera ID, will inherit the same ID of that previous node.
Otherwise, it will be assigned a new ID if there is no other previous node connected.
Therefore, we can obtain predicted tracklet IDs corresponding to every detection at the current frame to accomplish inference.

\section{Network Details}\label{sub:network}
\begin{figure}[t]
\begin{center}
\begin{subfigure}[b]{0.45\linewidth}
         \centering
         \includegraphics[width=1\linewidth]{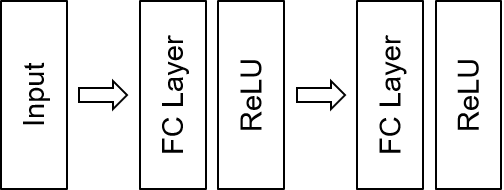}
         \caption{Encoder}
         \label{fig:mlp-a}
\end{subfigure}
\hfill
\begin{subfigure}[b]{0.45\linewidth}
         \centering
         \includegraphics[width=1\linewidth]{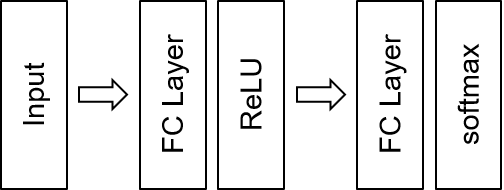}
         \caption{Classifier}
         \label{fig:mlp-b}
\end{subfigure}
\end{center}
   \caption{MLP network architecture. (a) $f^v_{FE}(\cdot)$, $f^e_{FE}(\cdot)$, $f^v_{ME}(\cdot)$, $f^e_{ME}(\cdot)$ have the same structure. (b) $f_{CLS}$ replaces ReLU with softmax to output a confidence score of each edge between 0 and 1.}
\label{fig:mlp}
\vspace{-0.5em}
\end{figure}

Following \cite{GNN-CCA}, our ReST model contains five trainable MLPs (Table \ref{tb:network}, Figure \ref{fig:mlp}).
$f^v_{FE}(\cdot)$ and $f^e_{FE}(\cdot)$ serve as initial feature encoders for nodes and edges to project the original features, e.g. appearance feature and geometry position, into a high-dimensional feature space.
$f^v_{ME}(\cdot)$ and $f^e_{ME}(\cdot)$ are used in MPN. We encode the feature first, and then pass the message and update it.
With the softmax layer appended at the end, $f_{CLS}(\cdot)$ outputs a confidence score between 0 and 1 from the enhanced edge feature.

Our graph model has about 154K parameters in total, which is a considerably light model compared with attention- or Transformer-based models \cite{TransMOT,DyGLIP}.
This makes ReST more suitable for real-world  application scenarios.

\begin{table}
  \begin{center}
  \resizebox{\columnwidth}{!}{%
  \begin{tabular}{c|c|c|c|c}
    \hline
    Network & Layer & Input & Output & Parameters \\
    \hline\hline
    \multirow{2}{*}{$f^v_{FE}(\cdot)$}
      & FC + ReLU & 512 / 514 & 128 & \multirow{2}{*}{69K / 70K} \\
      & FC + ReLU & 128 & 32 \\
    \hline
    \multirow{2}{*}{$f^e_{FE}(\cdot)$}
      & FC + ReLU & 4 / 6 & 8 & \multirow{2}{*}{94 / 110} \\
      & FC + ReLU & 8 & 6 \\
    \hline
    \multirow{2}{*}{$f^v_{ME}(\cdot)$}
      & FC + ReLU & 38 & 64  & \multirow{2}{*}{4576} \\
      & FC + ReLU & 64 & 32 \\
    \hline
    \multirow{2}{*}{$f^e_{ME}(\cdot)$}
      & FC + ReLU & 70 & 32 & \multirow{2}{*}{2470} \\
      & FC + ReLU & 32 & 6 \\
    \hline
    \multirow{2}{*}{$f_{CLS}(\cdot)$}
      & FC + ReLU & 6 & 4  & \multirow{2}{*}{33} \\
      & FC + softmax & 4 & 1 \\
    \hline
  \end{tabular}%
  }
  \end{center}
  \caption{Details of each MLP network. The number before slash represents spatial graph, while the number behind represents temporal graph.}
  \label{tb:network}
  \vspace{-0.5em}
\end{table}

\section{Analysis on Post-processing Module}\label{sub:pp}
To validate the effectiveness of our post-processing module, we perform another ablation study on the post-processing module. In Algorithm 2, both spatial and temporal graphs perform pruning and splitting, while assigning tracklet ID will only perform in the temporal graph. Pruning and assigning ID are necessary and cannot be omitted, since pruning removes the edges and divides into several connected components representing different objects, and assigning ID is to output tracklet ID for evaluation. In practice, splitting is performed in both graphs to ensure that each connected component follows the specific constraints. As shown in Table \ref{tb:post-processing}, there is a significant decline in both metrics without the splitting in any graph or in both graphs.

\begin{table}
  \begin{center}
  \resizebox{\columnwidth}{!}{%
  \begin{tabular}{c|c|c}
    \hline
    Setting & IDF1$\uparrow$ & MOTA$\uparrow$ \\
    \hline\hline
    w/o splitting in both graphs & 80.5 & 92.8 \\
    w/o splitting in spatial graph & 84.3 & 92.4 \\
    w/o splitting in temporal graph & 85.5 & 95.5 \\
    \hline
    Ours (w/ full post-processing) & \textbf{91.6} & \textbf{97.0} \\
    \hline
  \end{tabular}%
  }
  \end{center}
  \caption{Ablation of post-processing module on Wildtrack.}
  \label{tb:post-processing}
  \vspace{-0.5em}
\end{table}

\section{Graph Visualization}\label{sub:graph-vis}
\begin{figure}[t]
\begin{center}
\begin{subfigure}[b]{0.48\linewidth}
         \centering
         \includegraphics[width=1\linewidth]{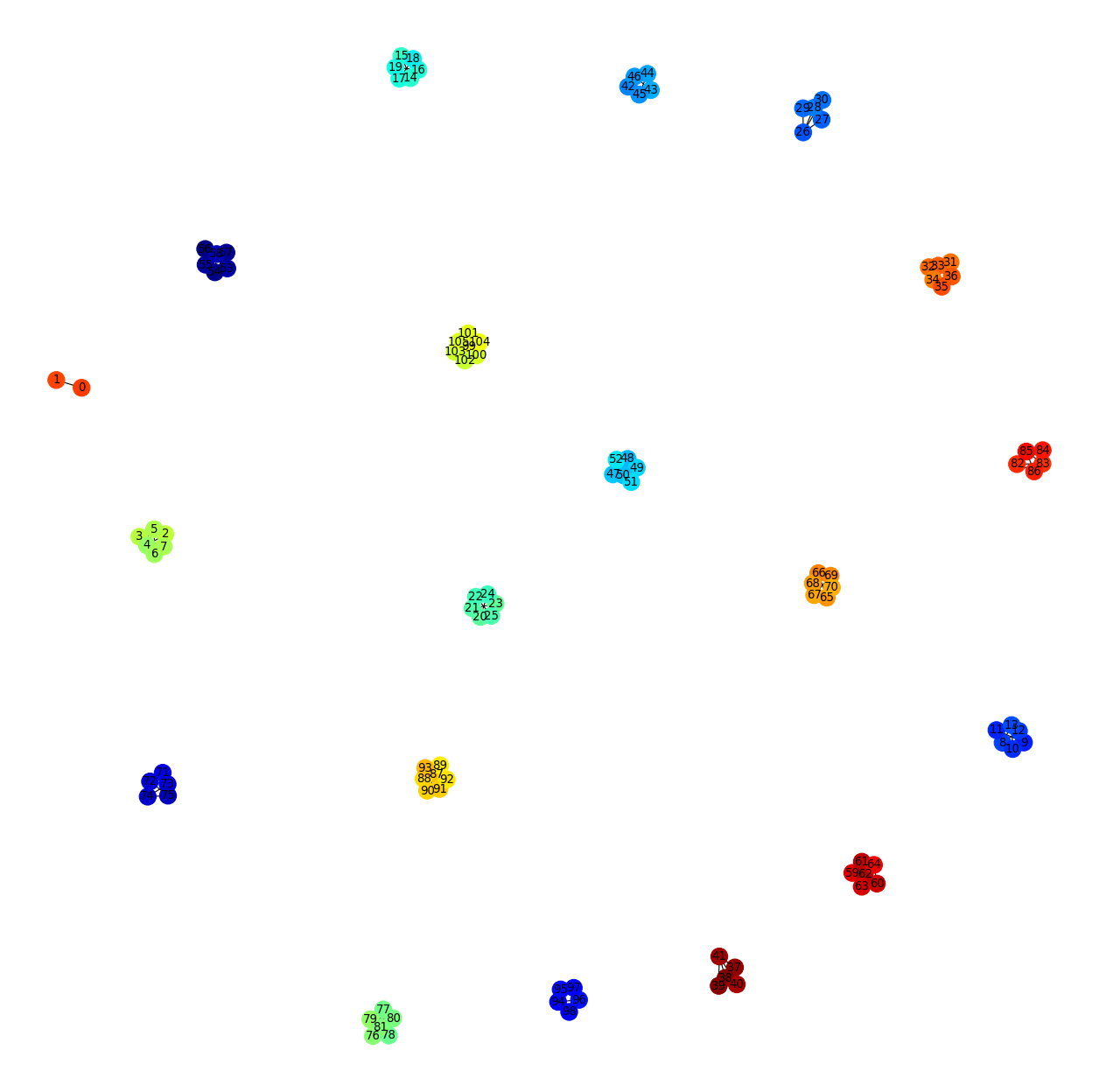}
         \caption{Spatial Graph}
         \label{fig:graph-a}
\end{subfigure}
\hfill
\begin{subfigure}[b]{0.48\linewidth}
         \centering
         \includegraphics[width=1\linewidth]{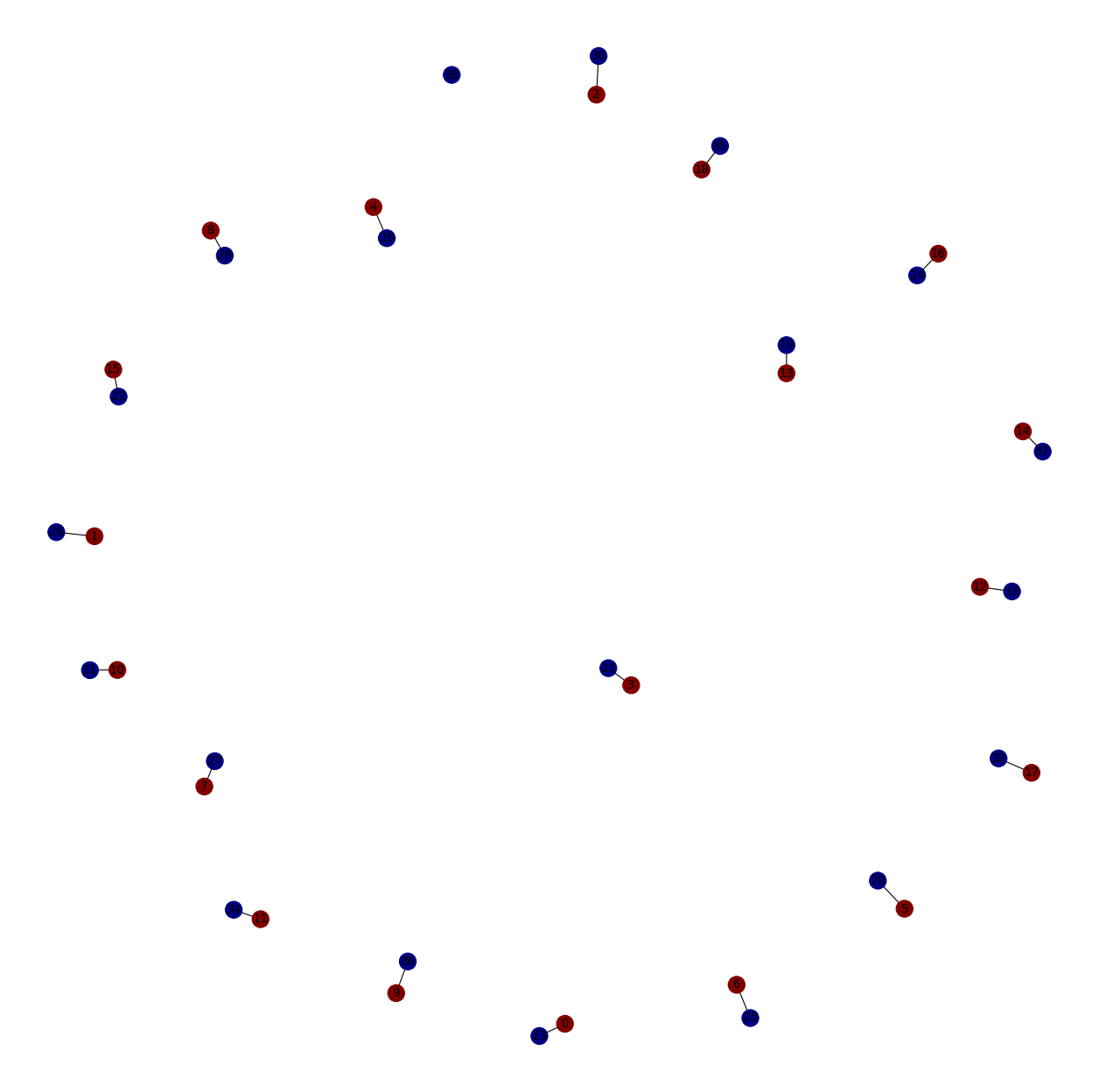}
         \caption{Temporal Graph}
         \label{fig:graph-b}
\end{subfigure}
\end{center}
   \caption{Graph visualization. Both graphs after their association stage are depicted. (a) In the spatial graph, each node represents one detection, colored by ground-truth label. (b) In the temporal graph, blue nodes represent aggregated nodes in temporal graph at previous frame, while red nodes represent aggregated nodes in spatial graph at current frame.}
\label{fig:graph}
\vspace{-1.5em}
\end{figure}

To better realize our graph model and prove the model robustness, we demonstrate the graph after the association stage in Figure \ref{fig:graph}. 
In Figure \ref{fig:graph-a}, one node, colored by ground-truth label, represents one input detction at current frame. Our spatial graph perfectly associates every people across different views even in a crowded scene. In other words, we will not lose the information of occluded people, leading to fewer fragmented tracklets and ID switch errors. 
In Figure \ref{fig:graph-b}, one node represents a list of detections that are aggregated in the Graph Reconfiguration stage. The connected nodes mean successful association between different frames, while the single nodes mean people who have just entered or left the scene. With the Graph Reconfiguration module, our view-invariant temporal graph becomes simple and focuses on associating nodes from different frames only.

\begin{figure*}
\begin{center}
\begin{subfigure}[b]{1\linewidth}
         \centering
         \includegraphics[width=1\linewidth]{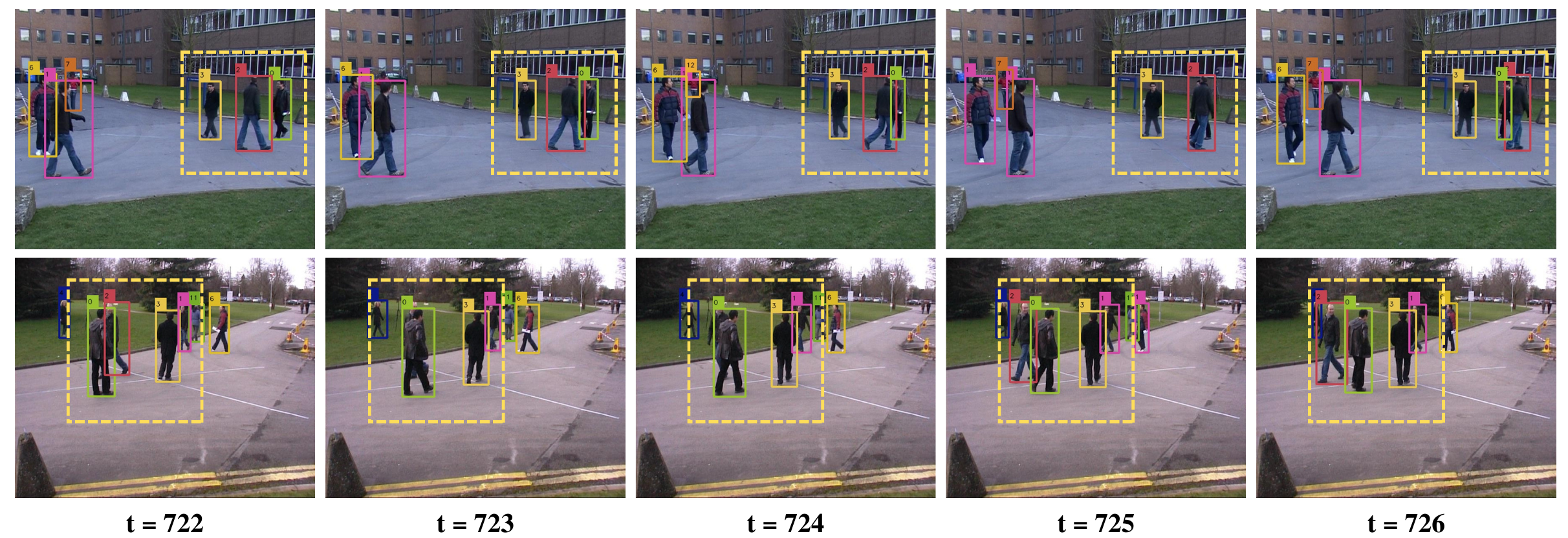}
         \caption{PETS09}
         \label{fig:qualitative-pets}
\end{subfigure}

\begin{subfigure}[b]{1\linewidth}
         \centering
         \includegraphics[width=1\linewidth]{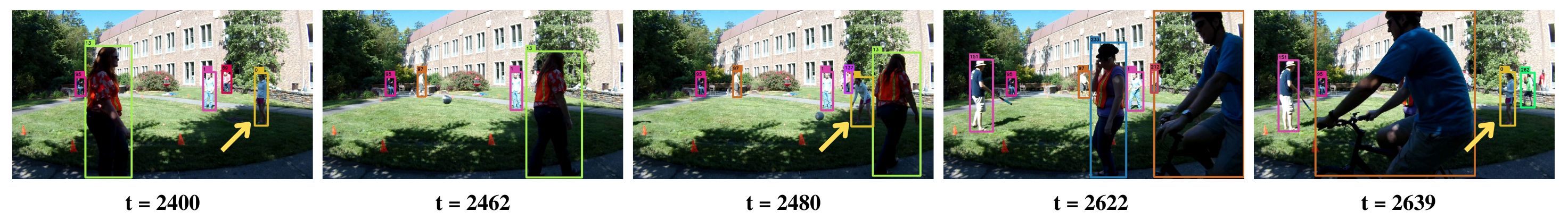}
         \caption{CAMPUS - Garden1}
         \label{fig:qualitative-campus}
\end{subfigure}
\end{center}
   \caption{Qualitative results on PETS09 \cite{PETS09} and CAMPUS \cite{HCT} to show our model's ability to recover fragmented tracklets. (a) ID:2 (red box) is occluded at time 723 to 724 in the second view, while he is still clearly visible in the first view. ID:0 (green box) is occluded at time 725 in the first view, but visible in the second view. Both cases are recovered and kept their IDs via our two-stages association. (b) Long-term consistency: ID:0 (yellow box) is occasionally occluded by people at time 2462 and 2622. Our model is able to maintain her ID in the long-term. As shown in the figure, we correct her ID at time 2480 and 2639.}
\label{fig:qualitative}
\end{figure*}


\section{Qualitative Results}\label{sub:qualitative}
In this section, we show more cases fixing the fragmented tracklets problem due to occlusion in certain views.
With the design of two-stages association, our ReST model leverages spatial consistency, recovering ID from different views.
Specifically, an occluded person is usually visible in other views.
We take advantage of this to fix the potential fragment and ID switch problems.
As shown in Figure \ref{fig:qualitative}, no matter the short-term or long-term occlusion, we steadily track every people and do not lose their tracklet IDs.


\section{Demonstration Video}\label{sub:demo}
We demonstrate our ReST tracker on Wildtrack \cite{wildtrack} at \url{https://github.com/chengche6230/ReST}.
To show spatial and temporal consistency, we simultaneously show all frames from the 7 camera views and a bird's-eye view of their foot point.
With the effective two-stages association, our predicted tracklet IDs are stable and consistent across views and frames.


\end{document}